\documentclass[sigconf]{acmart}
\usepackage{tikz}
\usepackage{comment}
\usepackage{amsmath} % define this before the line numbering.
\usepackage{color}
\usepackage{bbding}
\usepackage{pifont}
\usepackage{subfig}
\usepackage{pifont}
\usepackage{wasysym}
\usepackage{algpseudocode}
\usepackage{caption}
\usepackage[linesnumbered,ruled,vlined]{algorithm2e}
\SetKwInput{KwInput}{Input}                % Set the Input
\SetKwInput{KwOutput}{Output}
\newcommand{\VRD}{\textsc{ViReD\text{ }}}
\newcommand{\xmark}{\ding{55}}%

\AtBeginDocument{%
  }

\setcopyright{acmcopyright}
\copyrightyear{2018}
\acmYear{2018}
\acmDOI{XXXXXXX.XXXXXXX}

\setcopyright{none}
\renewcommand\footnotetextcopyrightpermission[1]{}

\acmConference[]{}{}
\begin{document}

%%
%% The "title" command has an optional parameter,
%% allowing the author to define a "short title" to be used in page headers.
\title{Learning to Retrieve Videos by Asking Questions}

%%
%% The "author" command and its associated commands are used to define
%% the authors and their affiliations.
%% Of note is the shared affiliation of the first two authors, and the
%% "authornote" and "authornotemark" commands
%% used to denote shared contribution to the research.
\author{Avinash Madasu}
\affiliation{%
  \institution{Department of Computer Science \\ UNC Chapel Hill}
  \country{USA}
}
\email{avinashm@cs.unc.edu}

\author{Junier Oliva}
\affiliation{%
  \institution{Department of Computer Science \\ UNC Chapel Hill}
  \country{USA}
}
\email{joliva@cs.unc.edu}

\author{
Gedas Bertasius}
\affiliation{%
  \institution{Department of Computer Science \\ UNC Chapel Hill}
  \country{USA}
}
\email{gedas@cs.unc.edu}

%%
%% By default, the full list of authors will be used in the page
%% headers. Often, this list is too long, and will overlap
%% other information printed in the page headers. This command allows
%% the author to define a more concise list
%% of authors' names for this purpose.
\renewcommand{\shortauthors}{Madasu et al.}

%%
%% The abstract is a short summary of the work to be presented in the
%% article.
\begin{abstract}
The majority of traditional text-to-video retrieval systems operate in static environments, i.e.,  there is no interaction between the user and the agent beyond the initial textual query provided by the user. This can be sub-optimal if the initial query has ambiguities, which would lead to many falsely retrieved videos. To overcome this limitation, we propose a novel framework for Video Retrieval using Dialog (\textsc{ViReD}), which enables the user to interact with an AI agent via multiple rounds of dialog, where the user refines retrieved results by answering questions generated by an AI agent. Our novel multimodal question generator learns to ask questions that maximize the subsequent video retrieval performance using (i) the video candidates retrieved during the last round of interaction with the user and (ii) the text-based dialog history documenting all previous interactions, to generate questions that incorporate both visual and linguistic cues relevant to video retrieval. Furthermore, to generate maximally informative questions, we propose an Information-Guided Supervision (IGS), which guides the question generator to ask questions that would boost subsequent video retrieval accuracy. We validate the effectiveness of our interactive \textsc{ViReD} framework on the AVSD dataset, showing that our interactive method performs significantly better than traditional non-interactive video retrieval systems. We also demonstrate that our proposed approach generalizes to the real-world settings that involve interactions with real humans, thus, demonstrating the robustness and generality of our framework.
\end{abstract}

%\vspace{-1.5cm}
%%
%% The code below is generated by the tool at http://dl.acm.org/ccs.cfm.
%% Please copy and paste the code instead of the example below.
%%


%%
%% Keywords. The author(s) should pick words that accurately describe
%% the work being presented. Separate the keywords with commas.
%% A "teaser" image appears between the author and affiliation
%% information and the body of the document, and typically spans the
%% page.

%%
%% This command processes the author and affiliation and title
%% information and builds the first part of the formatted document.
\maketitle

\begin{figure*}
\centering
\includegraphics[width=0.8\linewidth]{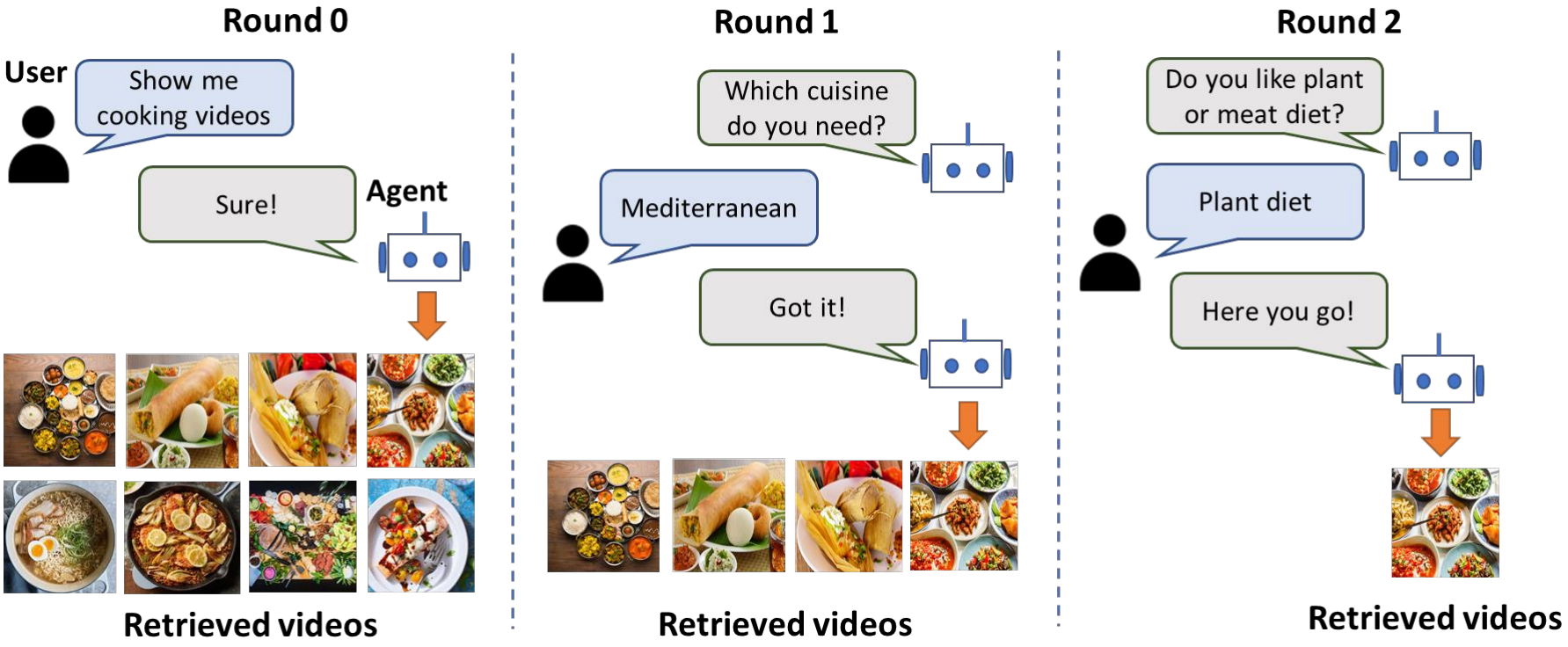}
\vspace{-0.2cm}
\caption{An illustration of our interactive dialog-based video retrieval framework. The order of conversation between the user and the agent is illustrated from left to right. The agent has access to a large video database, which is used for retrieving user-specified videos. For example, in this case, the user starts an interaction with the agent by asking it to ``Show some cooking videos.'' The agent then searches for relevant videos in the database and returns eight candidate videos. Due to high uncertainty in the initial query, the agent then asks another follow-up question ``Which cuisine do you prefer?'' for which the user responds: ``Mediterranean.'' As the number of retrieved video candidates is reduced to four, the agent asks one final question: ``Do you like plant or meat diet?'' The user's response (i.e., ``plant diet'') then helps the agent to reduce the search space to the final candidate video, which is then displayed to the user.\vspace{-0.2cm}}
\label{fig:motivation_diagram}
\end{figure*}
\section{Introduction}
The typical (static) video retrieval framework fetches a limited list of candidate videos from  a large collection of videos according to a user query (e.g.~`cooking videos'). However, the specificity of this query will likely be limited, and the uncertainty among candidate videos based on the user query is typically opaque (i.e.~the user might not know what additional information will yield better results). 

For example, consider the scenario where you are deciding what dish to make for dinner on a Friday night. Now also suppose that you have access to an interactive AI agent who can help you with this task by retrieving the videos of relevant dishes and detailed instructions on how to make those dishes. In this particular scenario, you might start an interaction with the agent by asking it to ``show some cooking videos'' (Figure \ref{fig:motivation_diagram} left-most query). Any traditional video retrieval model will look for the matching cooking videos from the database and display them to the user. However, what happens if there are too many matching videos, most of which don't satisfy the user's \emph{internal} criteria? A user-friendly video retrieval framework will not display all such videos to the user and expect them to sift through hundreds of videos to find the videos that are most relevant to them. Not only this would be overly time consuming, but it would also hurt the user experience. Instead, one way to address the uncertainty would be to ask another follow-up question of the same user: ``Which cuisine do you prefer?'' This would then allow the user to provide additional information clarifying some of his/her preferences (e.g., plant or meat diet, etc.) so that an AI agent can narrow down its search.

The rise of conversational AI systems, such as chat-bots and voice assistants, have made the user interaction with a digital agent relatively smooth. Inspired by this emerging technology, and a huge availability of video data, we propose \VRD, a framework for  \textbf{Vi}deo \textbf{Re}trieval using \textbf{D}ialog. Injecting dialog into standard text-to-video retrieval systems has two key advantages: (i) it reduces the uncertainty associated with the initial user text queries, (ii) it enables the agent to infer user's internal preferences, thus, making the AI model more personalized to the user. 

%To show how dialog improves the video retrieval, we compare the video retrieval performance with and without dialog as additional input. As shown in the figures \ref{fig:dialog_r1} and \ref{fig:dialog_r5}, using dialog significantly improves the performance. It demonstrates that dialog can improve video retrieval performance.

Several works \cite{flickner1995query, wu2004willhunter,rui1998relevance,kovashka2012whittlesearch,kovashka2013attribute} explored the idea of interactive mechanisms in the context of image retrieval. Prior methods in this area used relevance scores \cite{rui1998relevance,wu2004willhunter} and attribute comparisons \cite{kovashka2012whittlesearch,kovashka2013attribute} to get user feedback for retrieval. Additionally, the recent work of Cai \textit{et al.} \cite{cai2021ask} proposed Ask-and-Confirm, a framework that allows the user to confirm if the proposed object is present or absent in the image. One downside of these prior approaches is that they typically require many interaction rounds (e.g., $>5$), which increases user effort, and degrades user experience. Furthermore, these approaches significantly limit the form of the user-agent interaction, i.e., the users can only verify the presence or absence of a particular object/attribute in an image but nothing more. In contrast, our \VRD framework enables the user to interact with an agent using \emph{free-form} questions, which is a natural form of interaction for most humans. We also note that our interactive framework achieves excellent video retrieval results with a few (e.g., $2-3$) interaction rounds. 

Our key technical contribution is a multimodal question generator optimized with a novel Information-Guided Supervision (IGS). Unlike text-only question generators, our question generator operates on (i) the entire textual dialog history (if any) and (ii) previously retrieved top video candidates, which allows it to incorporate relevant visual and linguistic cues into the question generation process. Furthermore, our proposed IGS training objective enables our model to generate maximally informative questions, thus, leading to higher text-to-video retrieval accuracy. 

%We also note that unlike previous text-only based question generators, our model operates on (i) the entire textual dialogue history (if any) and (ii) previously retrieved top video candidates, which enables our model to incorporate relevant visual and linguistic cues into the question generation process. Furthermore, our question generator is trained using a novel Information-Guided Supervision (IGS) approach, which helps the model to generate maximally informative questions, thus, leading to higher subsequent text-to-video retrieval accuracy. 

%In comparison to these prior approaches, our \VRD framework addresses the above listed issues using our proposed multimodal question generator. 

We validate our entire interactive framework \VRD on the Audio-Visual Scene Aware Dialog dataset (AVSD)~\cite{alamri2019audio} demonstrating that it outperforms all non-interactive methods by a substantial margin. Furthermore, compared to other strong dialog-based baselines, our approach requires fewer dialog rounds to achieve similar or even better results. We also demonstrate that our approach generalizes to the real-world scenarios involving interactions with real humans, thus, indicating its effectiveness and generality. Lastly, we thoroughly ablate different design choices of our interactive video retrieval framework.

\vspace{-0.1cm}
\section{Related Work}
\subsection{Multimodal Conversational Agents}
There has been a significant progress in designing multimodal conversational agents especially in the context of image-based visual dialog~\cite{antol2015vqa, das2017visual, niu2019recursive, das2017learning, qi2020two}. Das \textit{et al.}\cite{das2017visual} proposed the task of visual dialog, in which an agent interacts with a user to answer questions about the visual video content. There also exists prior work in co-operative image guessing between a pair of AI agents~\cite{das2017learning}. Furthermore, the recent work of Niu \textit{et al.} \cite{niu2019recursive} proposes a recursive visual attention scheme for visual dialog generation. We note that most of these prior approaches operate in closed-set environments, i.e., selecting questions/answers from a small set of candidates. In contrast, our model leverages visual and linguistic cues to generate open-ended questions optimized for video retrieval. 

\vspace{-0.1cm}
\subsection{Video Question Answering}
Following standard visual question answering (VQA) methods in images \cite{teney2018tips, agrawal2016analyzing, agrawal2018don, mishra2019ocr}, video based question answering (video QA) aims to answer questions about videos \cite{zeng2017leveraging, lei2018tvqa, lei2019tvqa+, yang2020bert}. Compared to visual question answering in images, video question answering is more challenging because it requires complex temporal reasoning. Le \textit{et al.} \cite{le2019multimodal} introduced a multi-modal transformer model for video QA to incorporate representations from different modalities. Additionally, Le \textit{et al.} \cite{le2020bist} proposed a bi-directional spatial temporal reasoning model to capture inter dependencies along spatial and temporal dimensions of videos. Recently, Lin \textit{et al.} \cite{lin2021vx2text} introduced Vx2Text, a multi-modal transformer-based generative network for video QA. Compared to these prior methods, we aim to develop a framework for interactive dialog-based video retrieval setting.

\subsection{Multimodal Video Retrieval}
Most of the recent multimodal video retrieval systems are based on deep neural networks \cite{dzabraev2021mdmmt, gabeur2020multi, chen2014multi, mithun2018learning, bain2021frozen, fang2021clip2video, croitoru2021teachtext}. With the advent of transformer-based language models \cite{radford2019language, lewis2019bart, raffel2019exploring}, several methods proposed transformer architectures for video retrieval~\cite{bain2021frozen, fang2021clip2video, croitoru2021teachtext}. However, these methods focus on static query-based video retrieval and perform poorly when the textual user queries are ambiguous. In contrast, our work proposes to use dialog as means to gain additional information for improving video retrieval results.

\subsection{Interactive Modeling Techniques}
Ishan \textit{et al.} \cite{misra2018learning} proposed an interactive learning framework for visual question answering. In this framework, the agent actively interacts with the oracle to get the information needed to answer visual questions. Several approaches utilized interactive mechanisms to perform image retrieval \cite{cai2021ask, wu2004willhunter, kovashka2012whittlesearch, murrugarra2021image, maeoki2020interactive}. Additionally, multiple interactive video retrieval methods were proposed for the  video browser showdown (VBS) benchmark~\cite{6945294}. These prior methods used interactive interfaces to obtain additional information related to the original search query from the user. This includes information such as attribute-like sketches~\cite{heller2022multi}, temporal context~\cite{lokovc2022video, ma2022reinforcement}, color~\cite{tran2022v}, spatial position~\cite{10.1007/978-3-030-98355-0_54}, and cues from the Kinect sensors~\cite{10.1145/2393347.2396432}. In comparison, we note that our proposed dialogue-based video retrieval framework is complementary to these prior approaches. In particular, we believe that combining dialog with the above listed cues (e.g., sketches, position, color, etc.) should boost the subsequent video retrieval accuracy. 

%\AM {\subsection{Video browser showdown}
%The video browser showdown (VBS) is a video retrieval competition in which the teams design interfaces for interactive feedback from the users \cite{heller2022multi, lokovc2022video, luu2022cdc, tran2022v, ma2022reinforcement}. The feedback from the users is collected through the attributes like sketches \cite{heller2022multi}, temporal context \cite{lokovc2022video, ma2022reinforcement} and color \cite{tran2022v}. Our proposed approach is complementary to these methods and combining dialog with these visual attributes can benefit the video retrieval accuracy.
%}

%We would also like to note that in general, we believe that our proposed dialogue-based video retrieval framework ViReD is complementary to many of the previous approaches applied to the VBS benchmark. In particular, we believe that combining dialog with the user-provided sketches, position and color information, inputs to the Kinetics sensors, or other interactive cues. should be useful for the subsequent video retrieval accuracy.

%Instead, we propose an interactive dialog-based framework for video retrieval. 

%Our work is some-what related to \cite{maeoki2020interactive} in terms of the problem statement. But, there is no interactive mechanism in \cite{maeoki2020interactive} to perform video retrieval.
%To the best of knowledge, we are the first to propose interactive dialogue-based framework for video retrieval. 

\begin{figure}
    \centering
    \includegraphics[height=4.3cm]{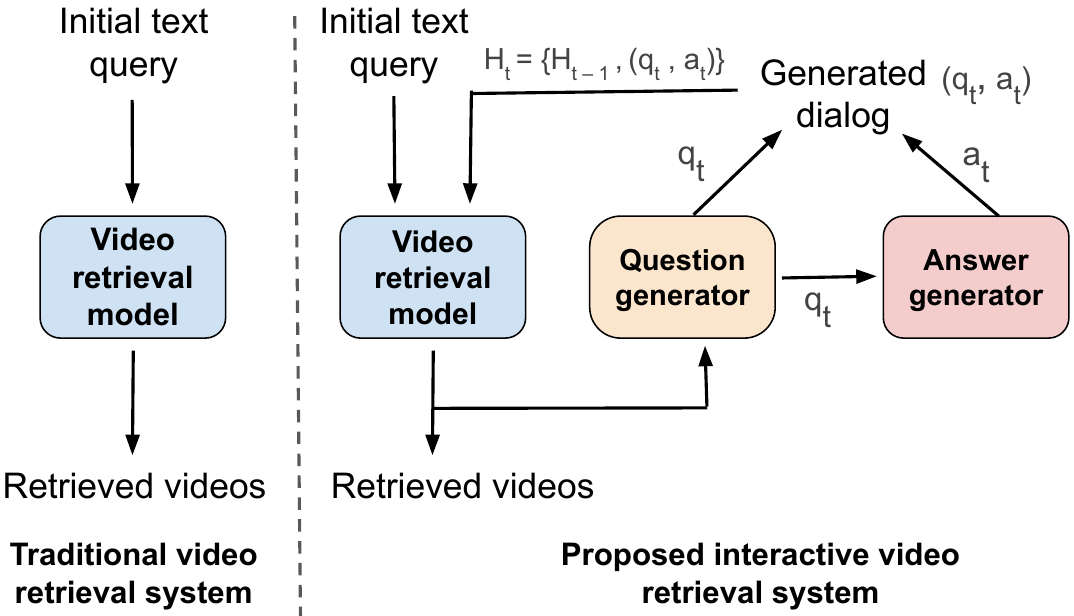}
    \vspace{-0.1cm}
    \caption{Comparison between the traditional (i.e., static) and our interactive dialog-based video retrieval frameworks. In the traditional set-up, the user interacts with the agent once by providing a single textual query to retrieve the desired video. In comparison, our proposed framework leverages multiple rounds of dialog with the user to improve video retrieval performance. Specifically, after the initial user query, the first round of retrieved videos are used to generate a question $q_{t}$, which the user then answers with an answer $a_{t}$. The generated dialog is added to the  dialog history $H_{t} = \{H_{t-1}, (q_{t}, a_{t})\}$, which is then used as additional input in the subsequent rounds of interaction.\vspace{-0.3cm}}
    \label{fig:proposed_approach}
\end{figure}

\begin{comment}
\begin{figure}
 \begin{subfigure}{0.15\textwidth}
   \includegraphics[width=\linewidth]{Figa.pdf}  
   \caption{R@1 metric}
   \label{fig:dialog_r1}
\end{subfigure}
 \begin{subfigure}{0.33\textwidth}
  \includegraphics[width=\linewidth]{Figb.pdf}  
   \caption{R@5 metric}
   \label{fig:dialog_r5}
 \end{subfigure}
 \caption{Comparison of video retrieval performance with and without dialog as an additional input.}
 \label{fig:dialog_effectiveness}
 \end{figure}
\end{comment}

\section{Video Retrieval using Dialog}
In this section, we introduce \VRD, our proposed video retrieval framework using dialog. Formally, given an initial text query $T$ specified by the user, and the previously generated dialog history $H_{t-1}$, our goal is to retrieve $k$ most relevant videos $V_{1}, V_{2}, ... , V_{k}$.

Our high-level framework, which is illustrated in Figure~\ref{fig:proposed_approach}, consists of three main components: (i) a multimodal question generator trained with an information-guided supervision (IGS), (ii) an answer generation oracle, which can answer any questions about a given video, thus, simulating an interaction with a human, and (iii) a video retrieval module, which takes as inputs the initial textual query and any generated dialog history and retrieves relevant videos from a large video database. We now describe each of these components in more detail.

%Different stages of the proposed approach is shown in the figure.

%\subsection{Initial set-up}

%This set up is shown in figure \ref{fig:inital_ret_setup}.
%\subsection{Interactive mechanism}
%After the initial set-up, the interactive mechanism is activated. 
%Our proposed interactive mechanism has two components. (i) question generator (ii) answer generator. 

\begin{figure}[!t]
\centering
\includegraphics[height=8.6cm]{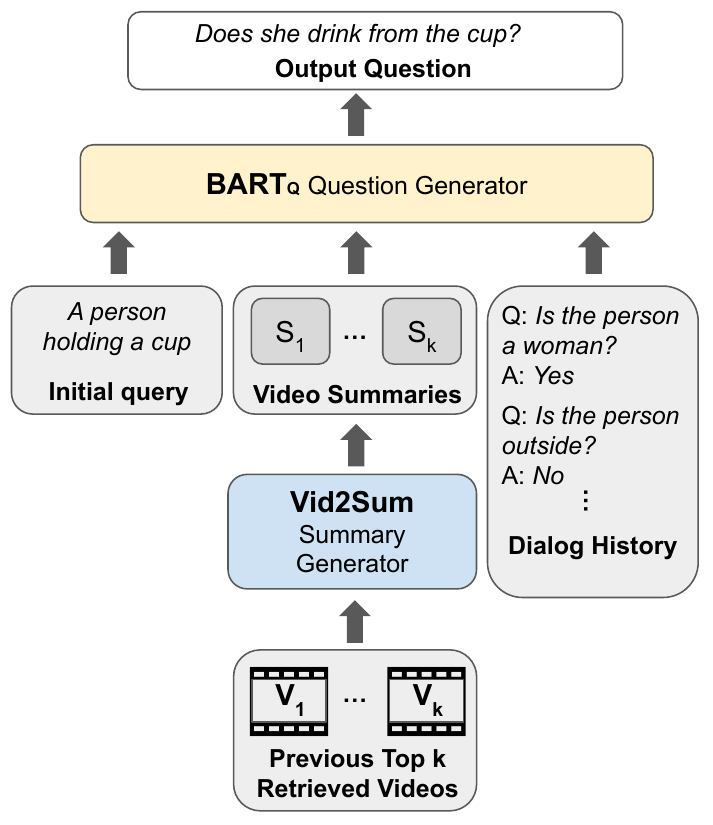}
\vspace{-0.2cm}
\caption{Illustration of the proposed question generator. It receives (i) an initial user-specified textual query,  (ii)  top-k retrieved candidate videos (from the previous interaction rounds), and (iii) the entire dialog history as its inputs. We then use a pretrained caption generator (Vid2Sum~\cite{song2021towards}) to map the videos into text. Afterward, all of the text-based inputs (including the predicted video captions) are fed into an autoregressive BART model for a new question generation.\vspace{-0.3cm}}
\label{fig:question_generator}
\end{figure}

\subsection{Question Generator} \label{questiongen}

As illustrated in Figure \ref{fig:question_generator}, at time $t$, our question generator takes as inputs (i) the initial text query $T$, (ii) top $k$ retrieved videos at time $t-1$, and (iii) previously generated dialog history $H_{t-1}$. To eliminate the need for ad-hoc video-and-text fusion modules~\cite{lin2021vx2text, li2020hero}, we use Vid2Sum video caption model~\cite{song2021towards} trained on the AVSD dataset to predict textual descriptions for each of the top-$k$ previously retrieved videos. Specifically, given a video $V_i$, the Vid2Sum model provides a detailed textual summary of the video content, which we denote as $S_i$. Afterward, the predicted summaries for all $k$ videos retrieved at timestep $t-1$, denoted as $S_{1}, S_{2}, ..., S_{k}$, are fed into the question generator along with the initial textual query $T$ and previous dialog history $H_{t-1}$. More precisely, we concatenate the (i) initial text query, (ii) the predicted summaries and (iii) the previous generated dialog history and pass it to an autoregressive BART language model~\cite{lewis2019bart} for generating the next question.

\begin{equation}
    X_q = \textrm{Concat}(T, S_{1}, S_{2}, .. , S_{k}, H_{t-1}),
\end{equation}
\begin{equation}
    q_{t} = \textrm{BART$_q(X_q)$}.
\end{equation}

\subsection{Answer Generation Oracle} \label{answergen}

The answer generator serves as an oracle that can answer any questions about a given video. We design our answer generator with a goal of simulating the presence of a human in an interactive dialog setting. Our goal is to use our answer generator to answer any open-ended questions posed by our previously described question generator. This characteristic is highly appealing as it makes our framework flexible and applicable to many diverse dialog scenarios. In contrast, the majority of prior methods~\cite{cai2021ask} are typically constrained to a small set of closed-set question/answer pairs, which makes it difficult to generalize them to diverse real-world dialog scenarios. In our experimental section \ref{qualitative}, we also conduct a user-study evaluation demonstrating that our answer generation oracle effectively replaces a human answering the questions.

Our answer generator takes a video $V_{i}$ and a question $q_{t}$ as its inputs. Similar to before, we first use a pretrained video caption model Vid2Sum \cite{song2021towards} to output a detailed textual summary $S_i$, which we then use as part of the inputs to the answer generator. Afterward, the generated summary $S_{i}$ and the question $q_{t}$ are concatenated and passed to a separate BART answer generation model to  generate an answer $a_{t}$ about the video $V_{i}$. 
\begin{equation}
    X_a = \textrm{Concat}(S_{i}, q_{t}),
\end{equation}
\begin{equation}
    a_{t} = \textrm{BART$_a(X_a)$}.
\end{equation}

Note that the BART models used for question and answer generation have the same architecture but that their weights are different (i.e., they are trained for two different tasks).

With these individual components in place, we can now generate $t$ rounds of dialog using our previously defined question and answer generators. The whole dialogue history generated over $t$ rounds can then be written as:
\begin{equation}
    H_t = (\{q_{1}, a_{1}\}, \{q_{2}, a_{2}\}, .... , \{q_{t}, a_{t}\}).
\end{equation}
The generated dialog history $H_t$ is then used as an additional input to the video retrieval framework, which we describe below.

\subsection{Text-to-Video Retrieval Model} \label{video-text-framework}

% Our video retrieval model (VRM) takes an initial textual query 
% $T^{(j)}$ and previous dialog history $H^{(j)}_t$ and returns a  probability distribution $p^{(j)} \in \mathbb{R}^{N}$ that encodes the similarity between each video $V^{(i)}$ in the database of $N$ videos and the concatenated text query $[T^{(j)}, H^{(j)}_t]$. Formally, we can write this operation as:

% \begin{equation}
% \label{eq:vrm}
% \begin{aligned}
% p^{(j)} = \mathrm{VRM}(T^{(j)}, H^{(j)}_t),
% \end{aligned}
% \end{equation}

% where each $p^{(j)}_i$ value encodes a probability that a video $V^{(i)}$ is the correct video associated with the concatenated textual query $[T^{(j)}, H^{(j)}_t]$.

Our video retrieval model (VRM) takes an initial textual query 
$T$ and previous dialog history $H_t$ and returns a  probability distribution $p \in \mathbb{R}^{N}$ that encodes the (normalized) similarity between each video $V^{(i)}$ in the database of $N$ videos and the concatenated text query $[T, H_t]$. Formally, we can write this operation as:
\begin{equation}
\label{eq:vrm}
\begin{aligned}
p = \mathrm{VRM}(T, H_t),
\end{aligned}
\end{equation}
where $p_i$ encodes a probability that the $i^\mathrm{th}$ video $V^{(i)}$ is the correct video associated with the concatenated textual query $[T, H_t]$.

Our video retrieval model consists of two main components: (i) a visual encoder $F(V; \theta_v)$ with learnable parameters $\theta_v$ and (ii) a textual encoder $G(T, H_t; \theta_t)$ with learnable parameters $\theta_t$. During training, we assume access to a manually labeled video retrieval dataset $\mathcal{X} = \{(V^{(1)}, T^{(1)}, H^{(1)}_t), \ldots , (V^{(N)}, T^{(N)}, H^{(N)}_t)\}$, where $T^{(i)}$ and $H^{(i)}_t$ depict textual queries and dialog histories associated with a video $V^{(i)}$ respectively. As our visual encoder, we use a video transformer encoder~\cite{gberta_2021_ICML} that computes a visual representation $f^{(i)} = F(V^{(i)}; \theta_v)$ where  $f^{(i)} \in \mathbb{R}^{d}$. As our textual encoder, we use DistilBERT~\cite{Sanh2019DistilBERTAD} to compute  a textual representation $g^{(i)} = G(T^{(i)}, H^{(i)}_t; \theta_t)$ where $g^{(i)}  \in \mathbb{R}^{d}$. We can jointly train the visual and textual encoders end-to-end by minimizing the sum of video-to-text and text-to-video matching losses as is done in~\cite{bain2021frozen}:

%similarities  using a contrastive video-to-text matching loss~\cite{bain2021frozen}. 

\begin{equation}
\label{eq:contrastive_v2t}
\begin{aligned}
\mathcal{L}_\mathrm{v2t} & =
-\frac{1}{B}\sum_{i=1}^B\log\frac{ \exp({f}^{(i)} \cdot g^{(i)})}{\sum_{j=1}^{B}\exp({f}^{(i)}\cdot g^{(j)})},\\
\end{aligned}
\end{equation}

\begin{equation}
\label{eq:contrastive_t2v}
\begin{aligned}
\mathcal{L}_\mathrm{t2v} & = 
-\frac{1}{B}\sum_{i=1}^B\log\frac{ \exp({g}^{(i)}\cdot  f^{(i)})}{\sum_{j=1}^{B}\exp({g}^{(i)}\cdot f^{(j)})}.\\
\end{aligned}
\end{equation}

Here, $B$ is the batch size, and $f^{(i)}, g^{(j)}$ are the embeddings of the $i^{th}$ video and $j^{th}$ text embeddings (corresponding to the $j^{th}$ video) respectively. Note that we use the matching text-video pairs in a given batch as positive and all the other pairs as negative samples. %We use Frozen in Time (FiT) \cite{bain2021frozen} codebase to implement our text-to-video retrieval framework. 

During inference, given an initial user query $T$ and the previous dialog history $H_t$, we extract a textual embedding $g = G(T, H_t; \theta_t)$ using our trained textual encoder where $g \in \mathbb{R}^{1 \times d}$. Additionally, we also extract visual embeddings $f^{(i)} = F(V^{(i)};\theta_v)$ for every video $V^{(i)}$ where $i = 1 \ldots N$. We then stack the resulting visual embeddings $[f^{(1)}; \ldots ;f^{(N)}]$ into a single feature matrix $Y \in \mathbb{R}^{N \times d}$. Afterward, the video retrieval probability distribution $p \in \mathbb{R}^{1 \times N}$ is computed as a normalized dot product between a single textual embedding $g$ and all the visual embeddings $Y$. This can be written as: $p = \mathrm{Softmax}(gY^\top)$. For simplicity, throughout the remainder of the draft, we denote this whole operation as $p = \mathrm{VRM}(T, H_t)$.

%FiT is a multi-modal transformer-based video retrieval model pretrained on the large-scale WebVid-2M dataset. Formally, given a textual query $T$, dialog history and video $V$, the video retrieval model (VRM) is trained using a combination of text-video and video-text losses.

%\begin{equation}
%    L_{t2v} = 
%\end{equation}
%... %It projects the text and video representations onto the same embedding space and brings the representations closer if the text and video are similar. We use the pre-trained frozen in time text-to-video retrieval model as the base model. It is fine-tuned to retrieve videos using the initial text query $T$ and the generated dialog $H_{t}$. 
%\GB{This description is unclear and vague. Can you use the text-to-video and video-to-text matching equations from their paper to make this section more technical and more formal?}

\section{Information-Guided Supervision for Question Generation} 
% JO: I'm taking a stab at this section.
Our goal in the above-described question generation step is to generate questions that will maximize the subsequent video retrieval performance. To do so, the generator must be able to comprehend: (i) the information it has already obtained, e.g.~through the history of dialogue and initial query; (ii) its current belief and ambiguity over potential videos that should be retrieved to the user, e.g.~through the current top candidate videos; and (iii) the potential information gain of posing new questions, e.g.~the anticipated increase in performance that may be had by posing certain questions. 

Although providing currently known information and belief over videos is straightforward via the dialogue history and top-$k$ candidate videos, respectively, comprehending (and planning for) future informative questions is difficult. A major challenge stems from the free-form nature of questions that may be posed. There is a large multitude of valid next questions to pose. Thus, explicitly labeling the potential information gain of all valid next questions does not scale. One may define the task of posing informative queries as a Markov decision process (MDP), where the current state contains known information, actions include possible queries to make, and rewards are based on the number of queries that were made versus the accuracy of the resulting predictions \cite{jafa, gsmrl}. Previous  interactive image retrieval \cite{cai2021ask, murrugarra2021image} approaches have used similar MDPs optimized through reinforcement learning to train policies that may select next questions from a limited finite list. However, these reinforcement learning (RL) approaches suffer when the action space is large (as is the case with open-ended question generation) and when rewards are sparse (as is the case with accuracy after final prediction) \cite{gsmrl}. Thus, we propose an alternative approach, information-guided question generation supervision (IGS), that bypasses a difficult RL problem, by explicitly defining informative targets for the question generated based on a post-hoc search.

Suppose that for each video $V^{(i)}$, $i \in \{ 1, \ldots, N \}$, we also have $m$ distinct human-generated questions/answers relevant to the video $D^{(i)} = \{D^{(i)}_{1}, \ldots, D^{(i)}_{m}\}$. Typically, such data is collected independently to any particular video retrieval system;  e.g.~in the AVSD \cite{alamri2019audio} dataset, users ask (and answer) multiple questions about the content of a given video (without any particular goals in mind).
%such as  \textit{Is she a man or woman}, \textit{is there only one person?}.
However, these human-generated questions can serve as potential targets for our question generator. With IGS, we propose to filter through $D^{(i)}$ according to the retrospective performance as follows. During training, we collect targets for the question generator at each round of dialogue separately. Let $T^{(i)}$, be an initial textual query corresponding to ground truth video $V^{(i)}$. Then, also let $S_{t, 1}^{(i)}, \ldots, S_{t, k}^{(i)}$ be our predicted text summaries of top-$k$ retrieved candidate videos after the $t^\mathrm{th}$ rounds of dialogue, $H^{(i)}_t$ (note that $H^{(i)}_0 = \emptyset$). 
We try appending question/answers $(q,a)$ in $D^{(i)}$ not in $H^{(i)}_t$ to $H^{(i)}_t$ and see remaining question would most improve retrieval performance. That is, we collect

% \begin{equation}
% \begin{aligned}
% (q^{*(i)}_{t+1}, a^{*(i)}_{t+1}) = \underset{(q,a) \in (D^{(i)} \setminus H^{(i)}_t)}{\mathrm{argmin}} \ell(V^{(i)}, \mathrm{VRM}(T^{(i)}, H^{(i)}_t \cup \{(q, a)\}),
% \end{aligned}
% \end{equation}

\begin{equation}
\begin{aligned}
\begin{split}
(q^{*(i)}_{t+1}, a^{*(i)}_{t+1}) & = \underset{(q,a) \in (D^{(i)} \setminus H^{(i)}_t)}{\mathrm{argmax}} \left[\,\mathrm{VRM}(T^{(i)}, H^{(i)}_t \cup \{(q, a)\})\,\right]_i,\\
%& = \underset{(q,a) \in (D^{(i)} \setminus H^{(i)}_t)}{\mathrm{argmax}} p_i\\
\end{split}
\end{aligned}
\end{equation}
where $\mathrm{VRM}$ is our previously described video retrieval model (see Sec.~\ref{video-text-framework}). Note that here, $\left[\mathrm{VRM}(T^{(i)}, H^{(i)}_t \cup \{(q, a)\})\right]_i = p_i$, which depicts our previously defined retrieval probability between the ground truth video $V^{(i)}$, and the concatenated text query $T^{(i)}, H^{(i)}_t \cup \{(q, a)\}$.
Each of the retrospective best questions are then set up as a target for the question generator at the $t+1^\mathrm{th}$ round 

\begin{equation}
\begin{aligned}
\mathcal{D}_{t+1} = \left\{\left(\left(T^{(i)}, S_{t, 1}^{(i)}, \ldots, S_{t, k}^{(i)}, H^{(i)}_t\right),\, q^{*(i)}_{t+1} \right)\right\}_{i=1}^N,
\end{aligned}
\end{equation}
where $\left(T^{(i)}, S_{t, 1}^{(i)}, \ldots, S_{t, k}^{(i)}, H^{(i)}_t\right)$ are the respective initial query, our predicted text summaries of top-k previous retrievals, and dialogue history that are inputs to the question generator, $\mathrm{BART}_q$. The target question/answers are appended to the histories $H_{t+1}^{(i)} = H_{t}^{(i)} \cup \{(q^{*(i)}_{t+1}, a^{*(i)}_{t+1})\}$, and the next round of target questions $\mathcal{D}_{t+2}$ is similarly collected.
Please note that $\mathcal{D}_{t+1}$ depends on $\mathcal{D}_{t}$ since we consider appending questions to previous histories. That is, at each round we look for informative questions based on the histories seen at that round. Jointly, the dataset $\mathcal{D}_{1} \cup \mathcal{D}_{2} \cup \ldots \cup \mathcal{D}_{M}$ serve as a \emph{supervised} dataset to directly train the question generator, $\mathrm{BART}_q$, to generate informative questions.

\setlength{\tabcolsep}{4pt}
\begin{table*}[t]
\begin{center}
\caption{Comparison to prior video retrieval models. In the "Pretrain Data" column, we list external datasets used for pretraining. The "Dialog" and "Dialog Rounds" columns depict whether the dialog is used as additional input, and if so how many rounds of it. Based on these results, we observe that our \VRD approach outperforms all baselines, including a strong Frozen-in-Time baseline augmented with 10 rounds of human-generated ground truth dialog.}
\vspace{-0.3cm}
\label{table:videoretrieval}
\begin{tabular}{lllllllll}
\hline\noalign{\smallskip}
Method & Pretrain Data &  Dialog & Dialog Rounds ($\downarrow$) & R@1 ($\uparrow$) & R@5 ($\uparrow$) & R@10 ($\uparrow$) & MedianR ($\downarrow$) & MeanR ($\downarrow$)\\
\noalign{\smallskip}
\hline
\noalign{\smallskip}
LSTM \cite{maeoki2020interactive} & ImageNet~\cite{russakovsky2015imagenet} & \Checkmark
 & 10 & 4.2	& 13.5	& 22.1 & $-$	& 119 \\
FiT \cite{bain2021frozen} & WebVid2M~\cite{bain2021frozen} & \xmark & $-$ & 5.6 & 18.4 & 27.5 & 25& 95.4 \\ 
 FiT w/ Human dialog ~\cite{maeoki2020interactive}  & WebVid2M~\cite{bain2021frozen} & \Checkmark
 & 10 & 10.8 & 28.9 & 40 & 18 & 58.7 \\ \hline
\bf \VRD & WebVid2M~\cite{bain2021frozen} & \Checkmark &  \bf 3
 &12 & 30.5 & 42.1 &  17 & 69.1 \\
 \bf \VRD & CLIP~\cite{pmlr-v139-radford21a} & \Checkmark & \bf 3 & \bf 24.9 & \bf 49.0 & \bf 60.8 & \bf 6.0 & \bf 30.3 \\
\hline
\end{tabular}
\end{center}
\vspace{-0.2cm}
\end{table*}
\setlength{\tabcolsep}{4pt}

\section{Experiments} \label{experiments}
\subsection{Dataset} \label{dataset}
We test our model on the audio-visual scene aware dialog dataset (AVSD) \cite{alamri2019audio}, which contains ground truth dialog data for every video in the dataset. Specifically, each video in the AVSD dataset has $10$ rounds of human-generated questions and answers describing various details related to the video content (e.g., objects, actions, scenes, people, etc.). Thus, we believe that this dataset is well suited to our setting. In total, the AVSD dataset consists of $7,985$ training, $863$ validation, and $1,000$ testing videos \cite{maeoki2020interactive}. Throughout our experiments, we use standard training, validation and test splits.

\subsection{Implementation Details}

\subsubsection{Question Generator.}
%\hspace{\parindent} \textbf{Question Generator.}
We train our question generator using the BART large architecture. We set the maximum sentence length to 120. During generation, we use the beam search of size 10. The question generator is trained for 5 epochs with a batch size of 32.

\subsubsection{Answer Generator.}
%\textbf{Answer Generator.} 
We also use the BART large architecture to train our answer generator. Note that the question and answer generators use the same architecture but are trained with two different objectives, thus, resulting in two distinct models. The maximum sentence length for answer generation is set to 135. During generation, we use the beam search of size 8. The model is trained with a batch size of 32 for 2 epochs.

\subsubsection{Video Retrieval Model.}
%\textbf{Video Retrieval Model.}
We use Frozen-in-Time (FiT) \cite{bain2021frozen} codebase to implement our video retrieval model. Specifically, we fine-tune the provided pretrained model on the AVSD dataset for 20 epochs with a batch size of 16. Early stopping is applied if the validation loss doesn't improve for 10 epochs. We use AdamW \cite{loshchilov2017decoupled} optimizer with a learning rate of $3e^{-5}$.

%We fine-tune a pretrained Frozen-in-Time \cite{bain2021frozen} video retrieval model on the AVSD dataset. We train the model for 20 epochs with a batch size of 16. Early stopping is applied if the validation loss doesn't improve for 10 epochs. We use AdamW \cite{loshchilov2017decoupled} optimizer with a learning rate of $3e^{-5}$.

\subsubsection{Vid2Sum Captioning Model.}
We fine-tune the video captioning model on the training set of AVSD for 5 epochs. We use the same hyper parameters as in \cite{song2021towards}. During inference, we use our trained Vid2Sum model to generate textual summaries for each input video. The generated summary has a maximum length of 25.

\subsection{Evaluation Metrics}
We measure the video retrieval performance using standard Recall@k ($k = {1, 5, 10}$), and MedianR, MeanR evaluation metrics. Recall@k calculates the percentage of test data for which the ground-truth video is found in the retrieved $k$ videos (the higher the better). Additionally, the MeanR and MedianR metrics depict the mean and the median rank of the retrieved ground truth videos respectively (the lower the better). Unless noted otherwise, all models are averaged over 3 runs.

\subsection{Video Retrieval Baselines}
%We compare our proposed approach to the following baselines.

%\subsubsection{LSTM.}
%\hspace{\parindent} \textbf{LSTM~\cite{maeoki2020interactive}.} Maeok \textit{et al.} \cite{maeoki2020interactive} proposed an LSTM based architecture for interactive video retrieval. However, there is no interactive component in the model. The model is trained and tested using concatenated input of text query and ground truth dialog.

\hspace{\parindent} \textbf{LSTM~\cite{maeoki2020interactive}.} Maeok \textit{et al.} \cite{maeoki2020interactive} proposed an LSTM-based model that processes human-generated ground truth dialog for video retrieval. Unlike this prior approach, our interactive \VRD approach does not require ground truth dialog data during inference. Instead, during each round of interaction, our method generates novel open-ended questions that maximize video retrieval accuracy. %whereas the method in \cite{maeoki2020interactive} requires explicit ground truth dialog inputs during inference.

%\subsubsection{Frozen-in-Time.}
\textbf{Frozen-in-Time~\cite{bain2021frozen}.} We fine-tune the Frozen-in-Time (FiT) model to retrieve the correct video using the initial textual query $T$ as its input (without using dialog).

\textbf{Frozen-in-Time w/ Ground Truth Human Dialog.} We fine-tune the Frozen-in-Time model using the textual query and the full $10$ rounds of human-generated ground truth dialog history. Unlike our \VRD approach, which uses our previously introduced question and answer generators to generate dialog, this Frozen-in-Time w/ Dialog baseline uses $10$ rounds of manually annotated human dialog history during inference. In this setting, we concatenate $10$ rounds of ground truth dialog with the initial text query, and use the concatenated text for video retrieval.

%We note that this baseline is somewhat unfair to all other baselines as it uses ground truth 

%\subsubsection{Interactive Human Baseline.}
%We also compare the proposed approach to 10 rounds of ground truth dialog. The 10 rounds of ground truth dialog is concatenated with text query during training and testing.
%\subsection{Answer generation baselines}
%We also compare the proposed question generator the existing video question answering models.

\setlength{\tabcolsep}{8pt}
\begin{table*}
\begin{center}
\caption{Comparison to the state-of-the-art on the video question answering task on the AVSD dataset. Our answer generator outperforms most prior methods and achieves comparable performance to the state-of-the-art Vx2Text~\cite{lin2021vx2text} system.}
\vspace{-0.2cm}
\label{table:answergensota}
\begin{tabular}{lllllll}
\hline\noalign{\smallskip}
Model & BLEU-1 & BLEU-2 & BLEU-3 & BLEU-4 & METEOR & ROUGE-L\\
\noalign{\smallskip}
\hline
\noalign{\smallskip}
MA-VDS \cite{hori2019end} & 0.256 & 0.161 &	0.109 & 0.078 & 0.113 & 0.277 \\
QUALIFIER \cite{ye2022qualifier} & 0.276 & 0.177 & 0.121 & 0.086 & 0.119 & 0.294 \\
Simple \cite{schwartz2019simple} & 0.279 & 0.183 & 0.13 & 0.095 & 0.122 & 0.303 \\
RLM \cite{li2021bridging} & 0.289 & 0.198 & 0.145 & 0.11 & 0.14 & 0.337 \\
VX2TEXT \cite{lin2021vx2text} & \textbf{0.311} & \textbf{0.217} & \textbf{0.16} & \bf 0.123 & 0.148
 & 0.35\\
\bf Ours & 0.308 & 0.215 & 0.158 & 0.121 & \textbf{0.149}
 & \textbf{0.351} \\
\hline
\hline
\end{tabular}
\end{center}
\vspace{-0.2cm}
\end{table*}
%\vspace{-0.2cm}
\setlength{\tabcolsep}{4pt}

 \begin{figure}[t]
  \centering
  \includegraphics[width=0.72\linewidth]{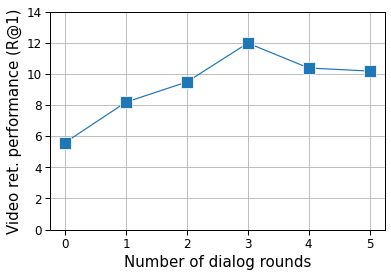}
  \vspace{-0.3cm}
  \caption{We study the video retrieval performance (R@1) as a function of the number of dialog rounds. Based on these results, we observe that the video retrieval accuracy consistently improves as we consider additional rounds of dialog. We also note that the performance of our interactive framework saturates after 3 rounds of dialog.\vspace{-0.4cm}}
\label{fig:dialog_turns_ablations}
\end{figure}

\begin{comment}
 \begin{figure}
\centering
\begin{subfigure}{\textwidth}
  \centering
  \includegraphics[width=\linewidth]{Qualitative-1.pdf}
  \caption{a}
  \label{fig:quali1}
\end{subfigure}%
\begin{subfigure}{\textwidth}
  \centering
  \includegraphics[width=\linewidth]{Qualitative-2.pdf}
  \caption{b}
  \label{fig:quali2}
\end{subfigure}
\caption{Visualization of the change in video ranking at each dialog turn. Ranking is ground truth video ranking.}
\label{fig:dialg_turns_ablations}
\end{figure}
\end{comment}

%\vspace{-0.1cm}
\section{Results and Discussion}

\subsection{Quantitative Video Retrieval Results}

In Table \ref{table:videoretrieval}, we compare our method with the previously described video retrieval baselines. We summarize our key results below.

%\subsubsection{Effectiveness of dialog}
%\textbf{The Importance of Pretraining.} 
\subsubsection{The Importance of Pretraining}
The results in Table \ref{table:videoretrieval} indicate that large-scale pretraining provides significant boost in video retrieval performance. Specifically, we note that the original Frozen-in-Time (FiT) baseline pretrained on large-scale WebVid2M~\cite{bain2021frozen} outperforms the previous state-of-the-art LSTM approach~\cite{maeoki2020interactive} by $1.4\%$ according to R@1 even without using any dialog data. Furthermore, we also note that using CLIP4Clip~\cite{luo2021clip4clip} backbone with CLIP~\cite{pmlr-v139-radford21a} pretraining leads to even better results, thus, indicating the importance of large-scale language-based pretraining.

%\subsubsection{Ablating the video retrieval models}
%In this ablation study, we replaced the frozen-in-time video retrieval model with the CLIP4Clip ~\cite{luo2021clip4clip} model and the results are shown in table \ref{table:videoretrieval}. CLIP4Clip significantly outperformed frozen-in-time model. 

\vspace{-0.1cm}
%\textbf{Dialog Effectiveness.} 
\subsubsection{Dialog Effectiveness}
Next, we demonstrate that dialog is a highly effective cue for the video retrieval task. Specifically, we first show that the FiT baseline augmented with $10$ rounds of human-generated ground truth dialog performs $5.2\%$ better in R@1 than the same FiT baseline that does not use dialog (Table \ref{table:videoretrieval}). This is a significant improvement that highlights the importance of additional information provided by dialog.

\vspace{-0.2cm}
%\textbf{The Number of Dialog Rounds.} 
\subsubsection{The Number of Dialog Rounds}
Next, we observe that despite using only $3$ rounds of dialog \VRD  outperforms the strong FiT w/ Human Dialog baseline, which uses $10$ rounds of ground truth dialog. It is worth noting that these $10$ rounds of  dialog were generated in a retrieval-agnostic manner (i.e., without any particular goal in mind), which may explain this result. Nevertheless, this result indicates that a few questions (e.g., $3$) generated by our model are as informative as $10$ task-agnostic human generated questions. 

In Figure \ref{fig:dialog_turns_ablations}, we also plot the R@1 video retrieval accuracy as a function of the number of the dialog rounds. We observe that the performance of our system consistently improves as we use more dialog rounds. Furthermore, we note that the performance saturates after $3$ rounds of interactions.

\setlength{\tabcolsep}{8pt}
\begin{table*}
\begin{center}
\caption{To validate the effectiveness of our interactive framework in the real-world setting, we replace our automatic answer generator oracle with several human subjects. Specifically, we randomly select 50 videos from the AVSD dataset, and ask 3 subjects to provide answers to our generated questions. We then use those dialogs to measure video retrieval performance as before. Our results suggest that our framework generalizes effectively to the settings involving real human subjects. The video retrieval is performed only on the subset of 50 selected videos.}
\vspace{-0.2cm}
\label{table:answergenhuman}
\begin{tabular}{llllll}
\hline\noalign{\smallskip}
Method & R@1 ($\uparrow$) & R@5 ($\uparrow$) & R@10 ($\uparrow$) & MedianR ($\downarrow$) & MeanR ($\downarrow$)\\
\noalign{\smallskip}
\hline
\noalign{\smallskip}
Answer Generator & 43.8 & 79.2 & 91.7 & 2.0 & 3.6 \\
Human Subject \#1 & 45.2 & 81 & 92.7 & 2.0 & 3.5\\
Human Subject \#2 & \bf 45.8 & \bf 81.2 & \bf 93 & \bf 2.0 & \bf 3.4\\
Human Subject \#3 &  45.6 & \bf 81.2 & 92.9 & \bf 2.0 & \bf 3.4 \\
\hline
\end{tabular}
\end{center}
\vspace{-0.2cm}
\end{table*}

\vspace{-0.2cm}
\subsection{Video Question Answering Results}

%\hspace{\parindent}
%\textbf{Comparison to the State-of-the-Art.} 
\subsubsection{Comparison to the State-of-the-Art}
As discussed above, we use our answer generator to simulate human presence in an interactive dialog setting. To validate the effectiveness of our answer generator, we evaluate its performance on the video question answering task on AVSD using the same setup as in Simple~\cite{schwartz2019simple}, and Vx2Text~\cite{lin2021vx2text}. We present these results in Table \ref{table:answergensota} where we compare our answer generation method with the existing video question answering baselines. Our results indicate that our answer generation model significantly outperforms many previous methods, including MA-VDS~\cite{hori2019end}, QUALIFIER~\cite{ye2022qualifier}, Simple~\cite{schwartz2019simple} and RLM~\cite{li2021bridging}. Furthermore, we note that our answer generator achieves similar performance as the recent Vx2Text model~\cite{lin2021vx2text}. These results indicate that our answer generator is comparable or even better than the state-of-the-art video-based question answering approaches.

\vspace{-0.1cm}
%\textbf{Replacing Our Answer Generator with a Human.} 
\subsubsection{Replacing Our Answer Generator with a Human Subject}

To validate whether our interactive framework generalizes to the real-world setting, we conduct a human study where we replace our proposed answer generator with several human subjects. To do this, we randomly select 50 test videos from AVSD, and ask $3$ human subjects to answer questions produced by our question generator. We then use the answers of each subject along with the generated questions as input to the video retrieval model (similar to our previously described setup). In Table \ref{table:answergenhuman}, we report these results for each of $3$ human subjects.  These results suggest that our interactive framework works reliably even with real human subjects. Furthermore, we note that compared to the variant that uses an automatic answer generator, the variant with a human in the loop performs only slightly better, thus, indicating the robustness of our automatic answer generation framework. Note that in this case, the video retrieval is performed on the subset of 50 selected videos.  %This result justifies our fully automatic framework where we replace the human in the loop with an automatic answer generator.

%human answers outperformed our answer generator. This is expected because human answers are more fine-grained compared to that of answer generator.

% \begin{figure}
%   \centering
%   \includegraphics[width=0.72\linewidth]{train_method_ablation.png}
%   \vspace{-0.2cm}
%   \caption{We study the effectiveness of using our proposed Information-guided Supervision (IGS) to train the question generator. As our baseline we train the question generator to generate questions in a video retrieval-agnostic fashion, i.e., using the same order as the human annotators did when they asked those questions. Our results indicate that our proposed IGS training objective produces superior performance compared to the video retrieval-agnostic baseline.\vspace{-0.3cm}}
%   \label{fig:train_methods_ablation}
%   \end{figure}

% \begin{figure}
% \centering
% \includegraphics[width=0.72\linewidth]{candidate_videos_ablation.png}
% \vspace{-0.2cm}
% \caption{We study the video retrieval performance in two settings: (i) when the question generator uses top-k retrieved videos as part of its inputs, and (ii) when it does not. In this case, k is set to 4. Based on the results, we observe that including top-k retrieved video candidates as part of the question generator inputs improves video retrieval accuracy for all number of dialog rounds.\vspace{-0.3cm}}
% \label{fig:candidate_videos_ablation}
% \end{figure}

\begin{figure}[t]
  \centering
  \includegraphics[width=0.72\linewidth]{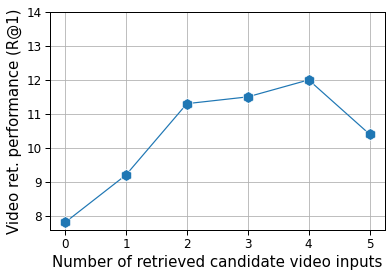}
  \vspace{-0.3cm}
\caption{We investigate the video retrieval performance as a function of the number of retrieved candidate video inputs that are fed into the question generator. These results indicate that the video retrieval performance is the best when we use 4 retrieved videos as inputs to our question generator.\vspace{-0.4cm}}
\label{fig:candidate_ablations}
\end{figure}

\setlength{\tabcolsep}{7pt}
\begin{table*}[t]
\begin{center}
\caption{We investigate the effectiveness of our question and answer generators with several different language models: BART-Base, T5-Small and T5-Large, and BART-Large. We evaluate our question generator with respect to the video retrieval task using using standard video retrieval metrics (the left part of the table). The results for the answer generator are evaluated using standard language generation metrics (see the right part of the table). Based on these results, we observe that BART-Large outperforms all the other variants according to all metrics. }
\vspace{-0.3cm}
\label{table:quesgenablation}
\begin{tabular}{l | lllll | llllll}
\hline\noalign{\smallskip}
\multicolumn{1}{c|}{Language} &  \multicolumn{5}{c|}{Question Generation for Video Retrieval} &  \multicolumn{6}{c}{Answer Generation}\\ 
 \hspace{0.3cm}Model & R@1 & R@5 & R@10 & MedianR & MeanR & BLEU-1 & BLEU-2 & BLEU-3 & BLEU-4 & METEOR & ROUGE-L \\
\noalign{\smallskip}
\hline
\noalign{\smallskip}
BART-Base&	8.4	     &   26.2	&37.4&	  22	&          87.5 & 0.292 & 0.197 &	0.141 & 0.107 & 0.132 & 0.321 \\
T5-Small	   &      8.9&	        27.1	&38.5&	  20&	         82.1 & 0.296 & 0.201 & 0.145 & 0.110 & 0.139 & 0.328 \\
T5-Large	&       11.2	&        29.4&	41.3&	17.5	& 72.3 & 0.303 & 0.207 &0.150 & 0.114 & 0.142 & 0.336 \\
BART-Large & \textbf{12} & \textbf{30.5} & \textbf{42.1} & \textbf{17} & \textbf{69.1} & \textbf{0.308} & \textbf{0.215} & \textbf{0.158} & \textbf{0.121} & \textbf{0.149}
 & \textbf{0.351} \\
\hline
\end{tabular}
\end{center}
\end{table*}
\setlength{\tabcolsep}{4pt}

\begin{figure*}[t]
\centering
\includegraphics[height=6.2cm]{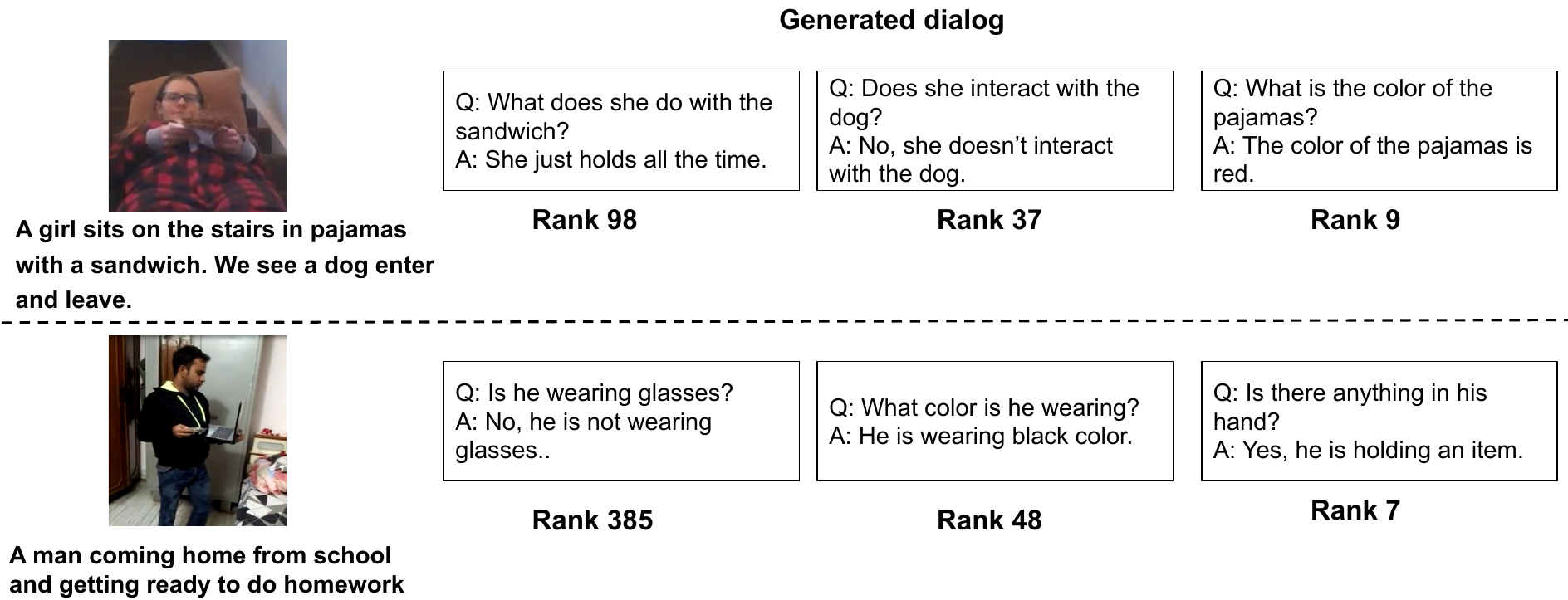}
\vspace{-0.1cm}
\caption{Qualitative results of our interactive video retrieval framework. On the left we illustrate the keyframe of the ground truth video $V_{gt}$ (i.e., the video that the user wants to retrieve) and the initial textual query for that video. From left to right, we visualize the three rounds of our generated dialog history (using our question generator and the answer generator oracle). Furthermore, under each dialog box, we also illustrate the rank of the ground truth video $V_{gt}$ among all videos in the database (i.e., the lower the better). Based on these results, we observe that each dialog round significantly improves video retrieval results (as indicated by the lower rank of the ground truth video). These results indicate the usefulness of dialog cues.\vspace{-0.0cm}}
\label{fig:quesgenvis}
\end{figure*}

\subsection{Ablation Studies} %% GB: stopped here

Next, we ablate various design choices of our model. Specifically, (i) we validate the effectiveness of our proposed Information-guided Supervision (IGS), (ii) the importance of using retrieved candidate videos for question generation, and (iii) how video retrieval performance changes as we vary the number of candidate video inputs to the question generator.

\subsubsection{Effectiveness of IGS}

To show the effectiveness of IGS, we compare the performance of our interactive video retrieval framework when using (i) IGS as a training objective to the question generator vs. (ii) using a video retrieval-agnostic objective. Specifically, we note the AVSD dataset has 
%10 rounds
10 pairs 
of questions and answers associated with each video. For the retrieval-agnostic baseline, we use the original order of the questions (i.e., as they appear in the dataset) to construct a supervisory signal for the question generator. In other words, we train our BART question generator to ask questions in the same order as the original AVSD human annotators did. In contrast, for our IGS-based objective, we order the questions such that they would maximize the subsequent video retrieval accuracy at each round of questions/answers. We report that IGS  outperforms the retrieval-agnostic baseline by $\bf 4.3\%$ in R\@1 video retrieval accuracy, which is a significant boost. Thus, this result validates the effectiveness of our proposed IGS technique.

%We illustrate these results in Figure~\ref{fig:train_methods_ablation} where we plot the video retrieval accuracy of both methods as a function of the number of dialog rounds. These results suggest that IGS significantly outperforms the retrieval-agnostic baseline, thus, validating the effectiveness of our proposed IGS technique.

%\subsubsubsection{Greedy order}
%This is a modification of IGS. In IGS, we measure the retrieval performance conditioned on previous informative questions. However, in greedy order we calculate the retrieval performance for each of the questions independently. The scores are sorted to determine the informative order of the questions. We train the question generator with the new order.

%We plot the video retrieval performance when the question generator is trained using original order, greedy order and IGS in the figure \ref{fig:train_methods_ablation}. From the figure \ref{fig:train_methods_ablation}, IGS significantly outperforms the original order and greedy order at all dialog turns. Greedy order outperforms the original order for dialog turns 1 and 2 but achieves similar performance as original order for 3 dialog turns. It shows that the greedy order resulted in the same set of questions as the original order in the third dialog turn which are not the most informative. This validates the effectiveness of IGS for training the question generator.

\subsubsection{The Importance of Using Retrieved Videos for Question Generation} \label{retvidablation}
%In Figure~\ref{fig:candidate_videos_ablation}, we verify the importance of using retrieved candidate videos for the question generation process. Specifically, we compare the video retrieval performance (i) when the question generator uses top-$k$ retrieved videos as part of its inputs and (ii) when it does not. Our results suggest that using top-$k$ retrieved videos for question generation produces a substantial boost in video retrieval performance. We use $k=4$ in this experiment.  %In the first scenario, the question generator generates questions conditioned on the initial text query and the previous generated dialog. In the second scenario, questions are generated conditioned on the retrieved videos, previous generated dialog and the initial text query. The performance comparison is shown in the figure \ref{fig:candidate_videos_ablation}. It is clear from the figure that the video retrieval performance is significantly higher when the questions are generated conditioned on the retrieved videos. 

We also verify the importance of using retrieved candidate videos for the question generation process. Specifically, we compare the video retrieval performance (i) when the question generator uses top-$k$ retrieved videos as part of its inputs and (ii) when it does not. Our results suggest that using top-$k$ retrieved videos for question generation produces a substantial boost of $\bf 3.9\%$ in R\@1 video retrieval performance. This indicates the usefulness of the additional video input modality to our question generator. We use $k=4$ in this experiment. 

\subsubsection{Ablating the Number of Video Inputs for Question Generation}
Next, in Figure~\ref{fig:candidate_ablations}, we study the video retrieval performance as a function of the number of retrieved video inputs fed to the question generator. These results indicate that the performance gradually increases with every additional video candidate input and reaches the peak when using $k=4$ retrieved videos. We also observe that the performance slightly drops if we set $k$ larger than $4$. We hypothesize that this happens because the input sequence length to the BART question generator becomes too long, potentially causing overfitting or other optimization-related issues.

%In the sub-section \ref{retvidablation}, we demonstrated that using retrieved videos for question generation significantly boosts the video retrieval performance. Now, we perform an ablation study to find the optimal number of retrieved videos that improve the performance of video retrieval. We plot the video retrieval as a function of the number of the retrieved videos and is shown in the figure \ref{fig:candidate_ablations}. We find that using the top-4 retrieved videos for question generation achieves the maximum performance on video retrieval.
% \begin{comment}
% \begin{figure}
% \centering
% \includegraphics[height=6cm]{Answer Generation.png}
% \caption{Illustration of the proposed answer generator. It receives caption, video and a question as the input. The video is converted to a caption using video to caption generator (Vid2Cap). The ground truth caption, generated caption and question are passed to BART model to generate answer to the input question.}
% \label{fig:answer_generator}
% \end{figure}
% \end{comment}

\subsubsection{Ablating the Choice of a Language Model}
We also investigate the effect of using different language models for our question and answer generators. Specifically, we experiment with BART-Base, T5-Small, T5-Large, and BART-Large models. The results are shown in Table \ref{table:quesgenablation}. From the results, it is evident that BART-large outperforms all the other models according to all evaluation metrics.

\subsection{Qualitative Results} \label{qualitative}

In Figure~\ref{fig:quesgenvis}, we also illustrate some of our qualitative interactive video retrieval results. On the left we show the keyframe of the ground truth video $V_{gt}$ (i.e., the video that the user wants to retrieve) and the initial textual query for that video. From left to right, we illustrate the three rounds of questions and answers produced by our question and answer generators. Additionally, under each question/answer box, we also visualize the rank of the video $V_{gt}$ among all videos in the database (i.e., the lower the better, where rank of $1$ implies that the correct video was retrieved). 

Our results indicate a few interesting trends. First, we observe that each dialog round boosts video retrieval performance, which is indicated by the lower rank of the ground truth video. Second, we note that our question generator learns to ask question that produce new pieces of information (i.e., information not mentioned in the initial textual query). Lastly, we observe that the questions asked by our model focus on diverse concepts including gender, presence of certain objects, human actions, clothes colors, etc. This highlights the generality of our open-ended question generator.

\begin{comment}
 \begin{figure}[!t]
\centering
\includegraphics[height=4cm]{Qualitative-1.pdf}
\caption{Visualization of the change in video ranking at each dialog turn. Ranking is ground truth video ranking.}
\label{fig:quesgenvis}
\end{figure}
\end{comment}

\begin{comment}
 \begin{figure}[!t]
 \centering
 \includegraphics[height=5cm]{open_ended_question_gen.png}
 \caption{Number of generated dialog }
 \label{fig:openendques}
 \end{figure}
\end{comment}

\section{Conclusions}

In this work, we introduced \VRD, an interactive framework for video retrieval using dialog. We demonstrated that dialog provides valuable cues for video retrieval, thus, leading to significantly better performance compared to the non-interactive baselines. Furthermore, we also showed that our proposed (i) multimodal question generator, and (ii) information-guided supervision techniques provide significant improvements to our model's performance. 

In summary, our method is (i) conceptually simple, (ii) it achieves state-of-the-art results on the interactive video retrieval task on the AVSD dataset, and (iii) it can generalize to the real-world settings involving human subjects. In the future, we will extend our framework to other video-and-language tasks such as interactive video question answering and interactive temporal moment localization.

\begin{acks}

This research was partly funded by NSF grant IIS2133595 and by NIH grant 1R01AA02687901A1.
\end{acks}

\begin{comment}
 \begin{figure}
\centering

\subfloat[a]{
	\label{subfig:correct}
	\includegraphics[width=0.45\textwidth]{Qualitative-1.pdf} } 

\subfloat[b]{
	\label{subfig:notwhitelight}
	\includegraphics[width=0.45\textwidth]{Qualitative-2.pdf} } 
\caption{Visualization of the change in video ranking at each dialog turn. Ranking is ground truth video ranking}
\label{fig:qual_results}
\end{figure}
\end{comment}
%%
%% The next two lines define the bibliography style to be used, and
%% the bibliography file.
\bibliographystyle{ACM-Reference-Format}
\bibliography{sample-base}

%%
%% If your work has an appendix, this is the place to put it.

\end{document}